\documentclass[conference]{IEEEtran}
\IEEEoverridecommandlockouts
\usepackage{cite}
\usepackage{amsmath,amssymb,amsfonts}
\usepackage{graphicx}
\usepackage{textcomp}
\usepackage{xcolor}
\usepackage{booktabs}
\usepackage{soul}

\usepackage{booktabs}
\usepackage{soul}

\usepackage{algorithm}
\usepackage[noend]{algpseudocode}
\usepackage{makecell}
\usepackage{multirow}

\usepackage[pagebackref,breaklinks,colorlinks]{hyperref}

\usepackage[capitalize]{cleveref}

\def\BibTeX{{\rm B\kern-.05em{\sc i\kern-.025em b}\kern-.08em
    T\kern-.1667em\lower.7ex\hbox{E}\kern-.125emX}}
\begin{document}

% \title{Unifying Global and Local Features for Improving End-to-End Temporally
% Precise Action Spotting in Sports Videos}

\title{Unifying Global and Local Scene Entities Modelling for Precise Action Spotting}

\author{
	Kim Hoang Tran$^{\ast\,1,3}$ \quad Phuc Vuong Do$^{\ast\,2,3}$ \quad Ngoc Quoc Ly$^3$ \quad Ngan Le$^4$ \\
	$^1$FPT Software AI Center, Vietnam \quad $^2$Naver Vietnam Dev Center, Vietnam \\
        $^3$VNUHCM-University of Science, Ho Chi Minh City, Vietnam \\
        $^4$Department of Computer Science, University of Arkansas, USA
}

\maketitle

\let\thefootnote\relax\footnote{$^{\ast}$These authors contributed equally to this work.}
\begin{abstract}
    Sports videos pose complex challenges, including cluttered backgrounds, camera angle changes, small action-representing objects, and imbalanced action class distribution. Existing methods for detecting actions in sports videos heavily rely on global features, utilizing a backbone network as a black box that encompasses the entire spatial frame. However, these approaches tend to overlook the nuances of the scene and struggle with detecting actions that occupy a small portion of the frame. In particular, they face difficulties when dealing with action classes involving small objects, such as balls or yellow/red cards in soccer, which only occupy a fraction of the screen space. To address these challenges, we introduce a novel approach that analyzes and models scene entities using an adaptive attention mechanism. Particularly, our model disentangles the scene content into the global environment feature and local relevant scene entities feature. To efficiently extract environmental features while considering temporal information with less computational cost, we propose the use of a 2D backbone network with a time-shift mechanism. To accurately capture relevant scene entities, we employ a Vision-Language model in conjunction with the adaptive attention mechanism. Our model has demonstrated outstanding performance, securing the 1st place in the SoccerNet-v2 Action Spotting, FineDiving, and FineGym challenge with a substantial performance improvement of 1.6, 2.0, and 1.3 points in avg-mAP compared to the runner-up methods. Furthermore, our approach offers interpretability capabilities in contrast to other deep learning models, which are often designed as black boxes. Our code and models are released at: \url{https://github.com/Fsoft-AIC/unifying-global-local-feature}.
\end{abstract}


\section{Introduction}

\begin{figure}[t]
    \centering
    \includegraphics[width=1.0\linewidth]{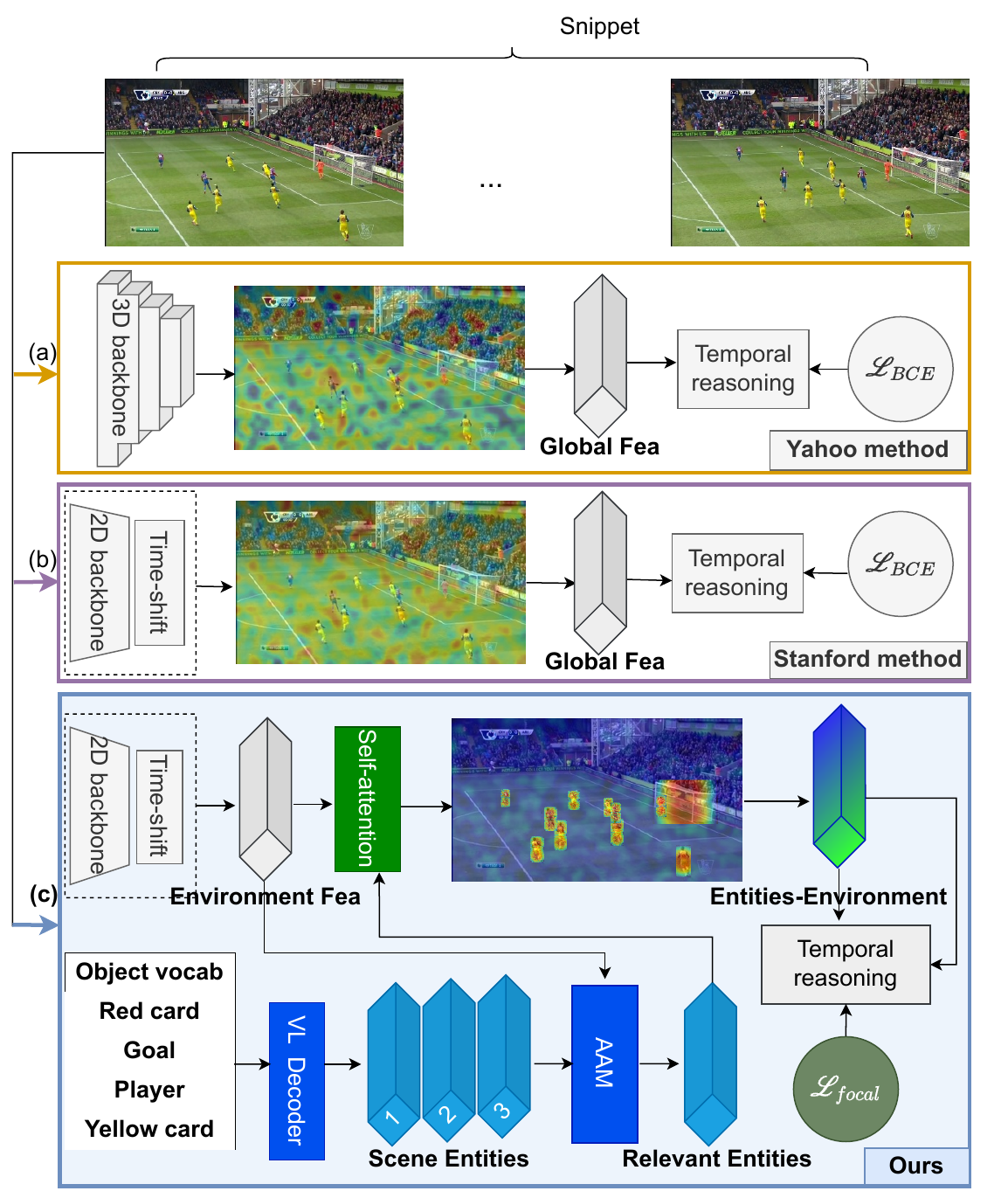}
    \caption{Comparison between our proposed framework with existing SOTA methods including the Yahoo method\cite{yahoo-soares-temporal} (a) and the Stanford method\cite{e2e_spotting} (b). The existing methods (a, b) extract features from the entire video frames, which may mistakenly overlook smaller key scene entities (e.g., red/yellow cards) that play a crucial role in actions. In contrast to these existing methods, our approach (c) disentangles the scene into global environmental features and local scene entities utilizing the Vision-Language (VL) model and an adaptive attention mechanism (AAM) to better focus on the pertinent entities that are actively involved in the action.}
    \label{fig:teaser}
\end{figure}
The emphasis on action in sports videos finds extensive use in broadcasting and sports analysis. However, efficiently identifying and highlighting actions within these videos can be resource-intensive and expensive. To alleviate the financial challenges associated with this task, several studies \cite{e2e_spotting, calf, netvlad, finediving, Shao2020FineGymAH, Chen2022FasterTADTT, Zhu2022ATS, calf-calib} have endeavored to tackle action spotting in sports videos. They aim to pinpoint specific moments at single timestamps that can streamline the automated creation of highlights. However, these endeavors have met with limited success due to their simplistic and ineffective methodologies, as well as the inherent challenges embedded in the datasets focused on action spotting in sports videos. The existing models for action spotting primarily consist of two main modules, a feature extractor and a temporal reasoning. While they excel in overall evaluation across all classes, they face challenges in predicting long-tail actions, or scenes containing small objects. Additionally, conspicuous challenges arise within sports videos: (i) scenes encompassing crowds, introducing background clutter, (ii) dynamic camera angle changes leading to varied perspectives, (iii) key sports elements such as ball, cards, goals, and referees occupying minimal frame space, and (iv) significant class-wise action imbalances, complicating data distribution.

% , which typically processes the entire spatial frame without focusing on the main objects contributing to the unfolding of the event.

% Recent advancements in the realm of sports video analysis, the cutting-edge E2E-spot \cite{e2e_spotting}, Yahoo\cite{yahoo-soares-temporal}, Baidu\cite{baidu-feature-combination-meet-attention} to accurately pinpointing events. 

To address these complexities, we propose enriching the action-spotting model with supplementary local information derived from the most relevant scene entities extracted via a Vision-Language (VL) model \cite{glip1} and Adaptive Attention Mechanism (AAM) \cite{vo2021aei, aoenet, yamazaki2023vltint}. Leveraging the recent advancements of VL in localizing objects, encompassing diverse scenarios such as small objects, diverse camera angles, and crowded backgrounds, we adopt Grounded language-image pre-trained model (GLIP) \cite{glip1} as our VL model. Particularly, we propose \underline{\textbf{U}}nifying \underline{\textbf{G}}lobal and \underline{\textbf{L}}ocal (UGL) module, which \emph{disentangles the scene content into global environment feature and local relevant scene entities feature}. To efficiently extract the global environment feature in both spatial and temporal dimensions with less computational cost, we utilize a 2D backbone network \cite{Alwassel2020TSPTP, Fan2021MultiscaleVT, regnet-y} and a time-shift module \cite{gsm}. To accurately capture the most relevant scene entities feature, we first utilize GLIP and RoiAlign methodology \cite{he2017mask} to extract all scene entities feature and we then employ AAM to select the most relevant scene entities feature.

It is worth noting that owing to its training-free nature and generalization capabilities, GLIP remains resilient against the imbalanced data challenges in sports videos. Specifically, our approach focuses on enhancing the network's performance when dealing with long-tail actions by placing a stronger emphasis on the relevant objects, even those that are small and directly involved in the action. In our proposed network, the global environmental features and the locally relevant scene entities are fused through a self-attention mechanism \cite{attention-is-all-you-need} during the creation of the entities-environment feature. Furthermore, to address the issue of imbalanced action class distribution in sports, we advocate training the network with Focal Loss \cite{focal-loss} instead of Cross Entropy, as employed in existing works. A visual comparison between our proposed network and existing state-of-the-art (SOTA) methods \cite{e2e_spotting, yahoo-soares-temporal} is presented in Figure \ref{fig:teaser}.

It is important to highlight that our network's ability to disentangle the scene content into global environment and local relevant scene entities provides it with interpretability capabilities. This sets it apart from other existing methods that utilize pre-trained backbone networks on the entire scene, often leading to challenges such as background clutter, overlooking small key action entities, imbalanced distribution, and more.
The contributions of our work can be summarized into three key aspects:
(i) We present a novel end-to-end method designed for efficient and accurate action spotting. 
(ii) We introduce the UGL module, which effectively disentangles the scene content into global environmental features and locally relevant scene entities features.
(iii) Our approach has achieved the highest ranking in the SoccerNet-v2 Action Spotting challenge \cite{soccernet-v2}, FineGym \cite{Shao2020FineGymAH}, FineFDiving \cite{finediving}. 
\section{Related works} 
\noindent\textbf{Action spotting.} 
% \hl{need to be rewritten for sport in general, not soccer only} 
The concept of action spotting was first presented in the SoccerNet \cite{soccernet-v1}. It involves identifying a specific momentary event, known as an \emph{action}, using a \emph{single timestamp}. This differs from temporal action localization, which is characterized by defining the starting and ending points of actions \cite{activity-net}. As proposed in \cite{soccernet-v1}, the performance of action-spotting is evaluated by loose average mAP metric. In this evaluation, a detection is considered correct if it falls within a certain time window around the true event, with a loose error tolerance ranging from 5 to 60 seconds, as defined in \cite{soccernet-v1}. A tighter error tolerance of 1 to 5 seconds is also used in the evaluation, as specified in \cite{soccernet-v2}. In general, we can categorize existing approaches into two-phase approaches and end-to-end approaches, as follows:

% Moreover, Soccernet \cite{soccernet-v1} evaluates the performance of action-spotting models by the measure of loose average mAP where detection is deemed correct if it occurs within some time-window around the true event, with a loose error tolerance (5–60 seconds) as in \cite{soccernet-v1} and tight error tolerance (1–5 seconds) as in \cite{soccernet-v2}. 
% Lastly, SoccerNet-v2 \cite{soccernet-v2} broadened the scope of action spotting by incorporating 17 classes and also proposed a more precise measure of tight average-mAP with a tight error tolerance (1–5 seconds), which is more challenging. 
% Several baselines (CALF \cite{calf}, NetVLAD \cite{netvlad, action-vlad}, MaxPool \cite{soccernet-v1}) utilize the pre-trained ResNet \cite{resnet} backbone for the feature extractor. Since the feature extraction module is not trained on SoccerNet-v2 so these features have very limited information on this dataset. Recent work E2E-spot\cite{e2e_spotting} introduced the task of spotting temporally precise, aiming to spot fine-grained events in videos and perform evaluation on various sports datasets. 

\noindent
\underline{\textit{Two-phase approaches}}. Baidu \cite{baidu-feature-combination-meet-attention} performs fine-tuning multiple action recognition models which consist of TPN \cite{tpn4action}, GTA \cite{gta}, VTN \cite{vtn}, irCSN \cite{csn}, and I3D-Slow \cite{i3d-slow} on snippets of videos. During the testing phase, they extract features from videos (clips of frames), rather than on a per-frame basis. Extracted features are fed into temporal detectors like NetVLAD++ \cite{netvladpp} or Transformer \cite{attention-is-all-you-need} to predict action spots. Yahoo \cite{yahoo-soares-temporal} is a standard two-phase approach. In the first phase, the video chunk is decoded, and feature vectors are generated to represent either individual frames or groups of frames. To enhance the quality of these features, they conducted experiments involving the fusion of resampled combination features \cite{baidu-feature-combination-meet-attention} with the ResNet features from Deliege et al. \cite{soccernet-v2}. In the second phase, they process fused features to produce the action predictions. Baidu \cite{baidu-feature-combination-meet-attention} and Yahoo \cite{yahoo-soares-temporal} are two-phase methods, combining pre-extracted features from multiple (5 to 6) heterogeneous, fine-tuned feature extractors in the first phase and proposing downstream architectures in the second phase and losses on those features. 

\noindent
\underline{\textit{End-to-End approach.}}
In contrast to the two-phase methods, E2E-spot \cite{e2e_spotting} takes a different approach by directly training a simple and compact end-to-end model. E2E-spot combines the RegNet-Y backbone \cite{regnet-y} with Gate-Shift Networks (GSM) \cite{gsm} for feature extraction and employs Gated Recurrent Unit (GRU) \cite{empirical-gru} for temporal feature extraction. Notably, \textit{our method also belongs to this category of end-to-end approaches.}

\noindent\textbf{Pre-trained Vision-Language (VL) Models}
Recent computer vision tasks are trained from VL models supervision, which has shown strong transfer ability in improving model generality and open-set recognition. CLIP \cite{clip} is one of the first works effectively learning visual representations by large amounts of raw image-text pairs. After being released, it has received a tremendous amount of attention. Some other VL models ALIGN \cite{jia2021scaling}, ViLD \cite{gu2022open}, RegionCLIP \cite{zhong2022regionclip}, GLIP \cite{glip1}, UniCL\cite{yang2022unified}, X-DETR \cite{cai2022x}, LSeg\cite{lseg}, DenseCLIP \cite{denseCLIP}, OpenSeg \cite{openseg}, GroundingDINO \cite{liu2023grounding} have also been proposed later to illustrate the great paradigm shift for various vision tasks. Based on case studies, we can categorize VL Pre-training models into (i) image classification e.g. CLIP, ALIGN, UniCL; (ii) object detection e.g. ViLD, RegionCLIP, X-DETR, GroundingDINO and (iii) image segmentation e.g. LSeg, OpenSeg. The first category is based on matching between images and language descriptions by bidirectional supervised contrastive learning (UniCL) or one-to-one mappings (CLIP, ALIGN). The second category contains two sub-tasks of localization and recognition. The third group involves pixel-level classification by adapting a pre-trained VL model. \textit{In this work, we utilize GLIP \cite{glip1} to localize scene entities.}

%Recent advancements in computer vision tasks involve the utilization of pre-trained models trained using visual-language (VL) supervision. Notably, CLIP \cite{clip} was one of the pioneering works that effectively learned visual linguistic representations through a large volume of raw image-text pairs. Various other VL models such as LSeg \cite{lseg}, DenseCLIP \cite{denseCLIP}, OpenSeg \cite{openseg}, MaskCLIP \cite{mask-clip}, and GLIP \cite{glip1} were introduced, illustrating a significant variation in the paradigm for diverse vision-related tasks. This motivated us to look for ways to apply these large VL pre-trained models to video understanding, and specifically in this paper, action spotting.

%------------------------------------------------------------------------
\section{Our proposed method}

Given an input video $\mathcal{V}=\{v_t\}_{t=1}^{N}$, where $N$ is the number of frames and a set of $K$ action classes, the goal is to predict the (sparse) set of frame indices when an action occurs, as well as the event's class 
$(t,\hat{\mathbf{y}}_{t})\in \{1,\cdots,N\} \times \{\textbf{c}_1,\cdots, \textbf{c}_K\}$.
% where $\hat{\mathbf{y}}_{t}^{i}$ is the $i$-th predicted action of frame $v_t$. 

Given a $\delta$-frame snippet, which consists of $\delta$ consecutive frames, our model outputs a sequence of $\delta$ class scores, each is represented $(K + 1)$-dimensional vector $\hat{\mathbf{y}}_t$ (including 1 'no-event' background class). Let $\phi(.)$ be an encoding function to extract the visual feature $f_t$ of a frame $v_t$; the $\delta$-frame snippet can be represented as $\mathcal{F}$ as follows:
\begin{equation}
        \mathcal{F} =\{f_t\}_{t=1}^{\delta}, \text{ where } f_i  =\phi(v_t)
\end{equation}

Different from the existing works \cite{baidu-feature-combination-meet-attention, yahoo-soares-temporal, e2e_spotting}, which simply define $\phi(.)$ as a pre-trained backbone network \cite{baidu-feature-combination-meet-attention, Alwassel2020TSPTP, Fan2021MultiscaleVT, regnet-y}) to extract global feature of the entire video frame, we model $\phi(.)$ by the proposed Unifying Global and Local (UGL) module. This module has the ability to globally capture environment feature of the entire scene while also locally extracting features from relevant scene entities. Given the feature sequence $\mathcal{F}$, our Long-Term Temporal Reasoning Module (LTRM) is responsible for aggregating long-term temporal information and subsequently making per-frame class score predictions. In the subsequent subsections, we begin by introducing the UGL module in \ref{subsec:UGL}. Next, we present the LTRM module in \ref{subsec:LTRM}, followed by a comprehensive explanation of our training process in \ref{subsec:training-method}.

\subsection{Unifying Global and Local (UGL) Module}
\label{subsec:UGL}
\begin{figure*}[t]
    \centering
    \includegraphics[width=\linewidth]{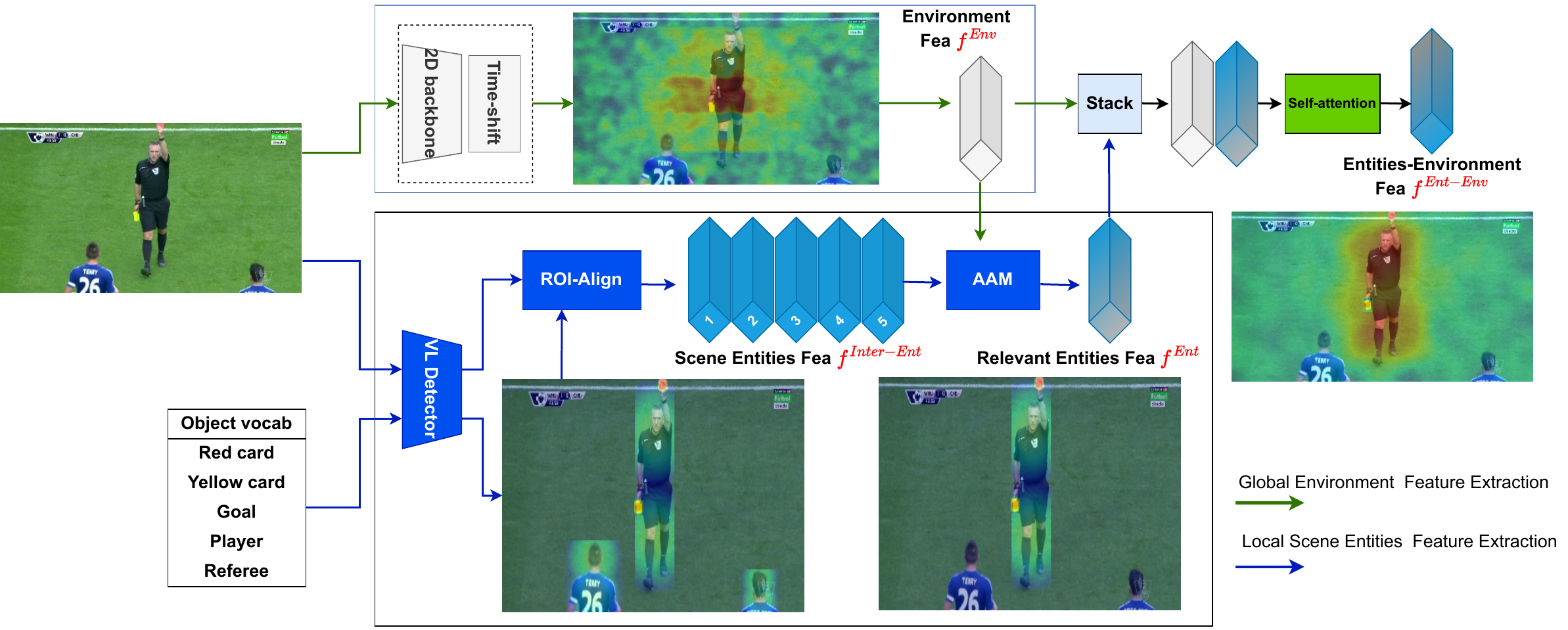}
    \caption{The architecture of our proposed UGL module. Given a frame $v_t$, our UGL module concurrently extracts both the global environment feature $f^{Env}$ and the local relevant entities feature $f^{Ent}$. Subsequently, it combines these features to produce a unified global-local entities-environment feature $f^{Ent-Env}$, acting as the unifying link between the local and global representations.
    }
    \label{fig:uglf}
\end{figure*}

Intuitively, UGL aims to extract features based on the principle of how a human perceives an action. This involves identifying the relevant scene entities during actions, and understanding the interactions between these relevant scene entities and their surrounding environment. In our work, we focus on exploring two modalities for extracting features: the global environment feature for capturing global context and the local relevant scene entities feature for highlighting relevant objects. UGL consists of three main components: (i) the extraction of global environment feature; (ii) the extraction of local relevant entities feature; and (iii) a fusion process that unifies the global-local entities-environment feature. The overall architecture of UGL is visualized in Fig.~\ref{fig:uglf}.

\subsubsection{Global Environment Feature}
This component plays a crucial role in capturing the global spatial information of an input frame $v_t$. To achieve this, we utilize a 2D-CNN network, RegNet-Y \cite{regnet-y}, as a backbone feature extractor for extracting a global-environment feature denoted as $f^{Env}$.  In comparison to other popular backbone networks such as VGG \cite{vgg}, ResNet \cite{resnet}, and EfficientNet \cite{efficient-net}, RegNet-Y stands out for its design efficiency and lower computing complexity. The RegNet-Y, as reported in \cite{regnet-y}, offers a smaller model size while still achieving comparable results. Consequently, RegNet-Y requires less time both in the training and inference stages.

Regarding incorporating temporal information, other works such as Baidu\cite{baidu-feature-combination-meet-attention}, and Yahoo\cite{yahoo-soares-temporal}, leverage the ensemble of 3D networks that are pre-trained on action recognition benchmark datasets to serve as a backbone feature extractor, and then they freeze this backbone. However, training an end-to-end model with such a backbone can be computationally intensive and costly. To address this challenge, we integrate the time-shift GSM (Gate-Shift Module) \cite{gsm} into the 2D backbone RegNet-Y. This effectively replaces the 3D backbone while enabling to capture the temporal information at a lower computational cost. In essence, the GSM module performs both forward and backward shifts in the spatial feature map along the time dimension, facilitating the sharing of temporal information across frames. As a result, we obtain the global environment feature $f^{Env}$ after applying RegNet-Y and GSM.

\subsubsection{Local Relevant Entities Feature}
Existing methods typically rely solely on the global environment features from pre-trained backbones and leave their model learning to spot action as a black-box. Moreover, they often face challenges related to imbalanced-class issues and may miss crucial information about relevant scene entities when scenes have intricate contexts and the sports relevant entities (e.g. yellow card, red card, ball) representing actions are particularly small. Therefore, it becomes essential to incorporate an object localization to semantically extract representations of the sports relevant entities.

\begin{figure*}[t]
    \centering
    \includegraphics[width=0.9\textwidth]{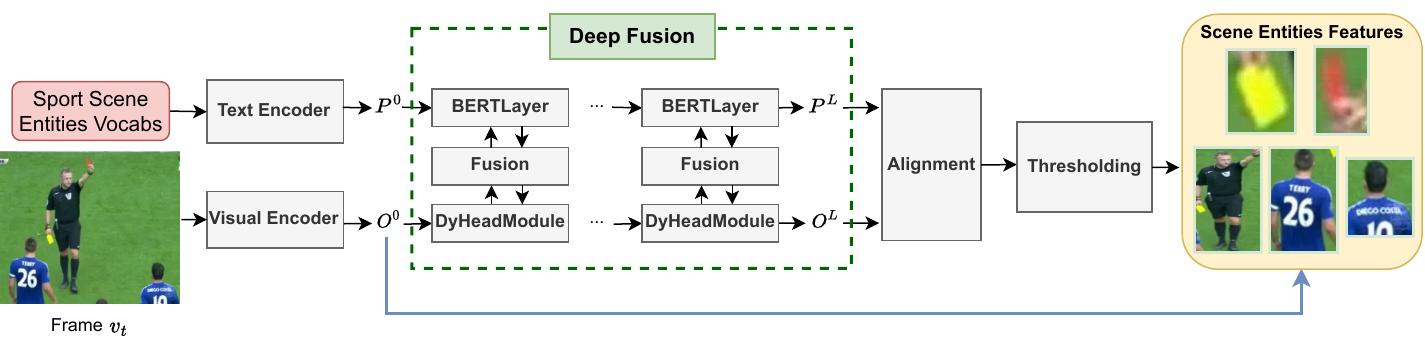}
    \caption{Pipeline to obtain the local scene entities features $f^{Inter-Ent}$ from GLIP, which correspond to the vocabulary of sports scene entities.}
    \label{fig:GLIP}
\end{figure*}

To eliminate the need of training the object detector on different datasets as in YOLO \cite{Redmon2015YouOL}, Faster-RCNN \cite{Ren2015FasterRT} as well as to detect a wide range of objects, we propose to leverage VL model GLIP \cite{glip1}. 
% , and detect diverse kinds of objects, we propose to use vision-language (V-L) object detection named GLIP \cite{glip1} instead of the traditional detectors like YOLO, RCNN. Moreover, the V-L detector GLIP can control detected objects based on the context of language, for example, it is able to correctly figure out "player on the field", while traditional detectors return all humans including the audience and player outside the field.
GLIP is a groundbreaking framework that unifies object detection and phrase grounding by redefining object detection as phrase grounding. It operates by taking a single image and a textual prompt that describes the objects to be detected in the task. Thanks to its utilization of a large VL pre-trained model, GLIP can efficiently identify most of the sports scene entities required without the need for additional training. The process of sports scene entities extraction is illustrated in \cref{fig:GLIP}. To be more specific, we start by defining a set of vocabulary related to sports scene entities (objects) and combine them in a text prompt. GLIP takes the input text prompt and processes them through a pre-trained language encoder, such as BERT \cite{bert}, to obtain a set of contextual word features denoted as $P^0 \in \mathbb{R}^{M \times d}$. Simultaneously, the input image undergoes processing via a visual encoder, which in our case is the Swin backbone \cite{swin-transformer}, resulting in a set of $N$ proposals represented as $O^0 \in \mathbb{R}^{N \times d}$. The deep fusion encoder is then employed to facilitate the matching of textual phrases with image proposals:

% It takes as input a single image and a text prompt describing the objects to be detected in the detection task. Due to the large vision-language pre-trained model, GLIP can yield most of the sports objects we need without training. The illustration of sports object extraction is shown in \cref{fig:GLIP}. Specifically, we pre-define a set of all vocabulary about sports objects and join them in a text prompt. GLIP extracts all word phrases from the input text prompt and passes them through a pre-trained language encoder (i.e., BERT \cite{bert}) to obtain $M$ contextual word features $P^0 \in \mathbb{R}^{M \times d}$. Meanwhile, the input image is processed through a visual encoder (i.e. Swin \cite{swin-transformer} backbone) to obtain a set of $N$ proposals $O^0 \in \mathbb{R}^{N \times d}$. The deep fusion encoder between visual embedding and text embedding is employed to match the textual phrases with the image proposals:
% \begin{equation}
%     O^i_{\text{t2i}}, P^i_{\text{i2t}} = \text{X-MHA}(O^i, P^i), \label{eq:xmha} \quad i \in \{0, 1, .., L-1\},
% \end{equation}
% \begin{equation}
%     O^{i+1} = \text{DyHeadModule}(O^i + O^i_{\text{t2i}}), \quad O = O^{L} \label{eq:i2t},
% \end{equation}
% \begin{equation}
%     P^{i+1} = \text{BERTLayer}(P^i + P^i_{\text{i2t}}), \quad P = P^{L} \label{eq:t2i} \\
% \end{equation}

\begin{align}
    & O^i_{\text{t2i}}, P^i_{\text{i2t}} = \text{X-MHA}(O^i, P^i), \label{eq:xmha} \quad i \in \{0, 1, .., L-1\}\\
    & O^{i+1} = \text{DyHeadModule}(O^i + O^i_{\text{t2i}}), \quad O = O^{L} \label{eq:i2t}, \\
    & P^{i+1} = \text{BERTLayer}(P^i + P^i_{\text{i2t}}), \quad P = P^{L} \label{eq:t2i},
\end{align}
where X-MHA in \cref{eq:xmha} represents the cross-modality multi-head attention module \cite{attention-is-all-you-need}, which computes the cross-attention between $O^i$ and $P^i$.

% After that, the word-region alignment module performs a dot product between the deep fused features to calculate the alignment score: 
After that, a dot product between the deep fused features is performed to calculate the alignment score: 
\begin{align}\label{eqn:ground_logits}
    & S^{\text{align}} = O P^{\top}, 
\end{align}
where $O = O^L \in \mathbb{R}^{N \times d}$ is a set of visual features from the last visual encoder layer and $P = P^L\in \mathbb{R}^{M \times d}$ is a set of word features from the last language encoder layer. The result of this operation are matrices $S^{\text{align}} \in \mathbb{R}^{N \times M}$. Each element $S^{\text{align}}(i,j)$ represents the alignment score between the $i$-th proposal scene entities (object) feature and the $j$-th word feature. To obtain the scene entities features of those detected boxes, we propose to use RoIAlign technique on the visual feature map $O^0$ extracted from the visual backbone, which is illustrated in \cref{fig:GLIP}.

Finally, we obtain a set of features for sports scene entities denoted as $f^{Inter-Ent}$. To effectively extract the most relevant scene entities $f^{Ent}$, from this set $f^{Inter-Ent}$, we employ an AAM \cite{aoenet}. In essence, AAM takes both the local scene entities features $f^{Inter-Ent}$ and the global environment feature $f^{Env}$ into consideration. Within the AAM framework, a hard attention mechanism \cite{adaptive-hard-attention} is initially utilized, followed by a soft self-attention mechanism \cite{attention-is-all-you-need}. These mechanisms aim to select and fuse all the most relevant scene entities into a single local relevant scene entities representation, $f^{Ent}$.

\subsubsection{Fusion Component}
The primary objective of this component is to capture the relationships between the global environment feature $f^{Env}$ and the local relevant scene entities feature $f^{Ent}$. To achieve this, we begin by concatenating these two types of features together. Subsequently, we apply a self-attention model \cite{attention-is-all-you-need}, followed by an average pooling layer, to combine the concatenated features into a unified representation denoted as $f^{Ent-Env}$. This representation encapsulates both the global environment feature and the local relevant scene entities, effectively modeling their relationship.

\subsection{Long-term Temporal Reasoning (LTR) Module}
\label{subsec:LTRM}
\begin{figure*}
    \centering
    \includegraphics[width=0.7\linewidth]{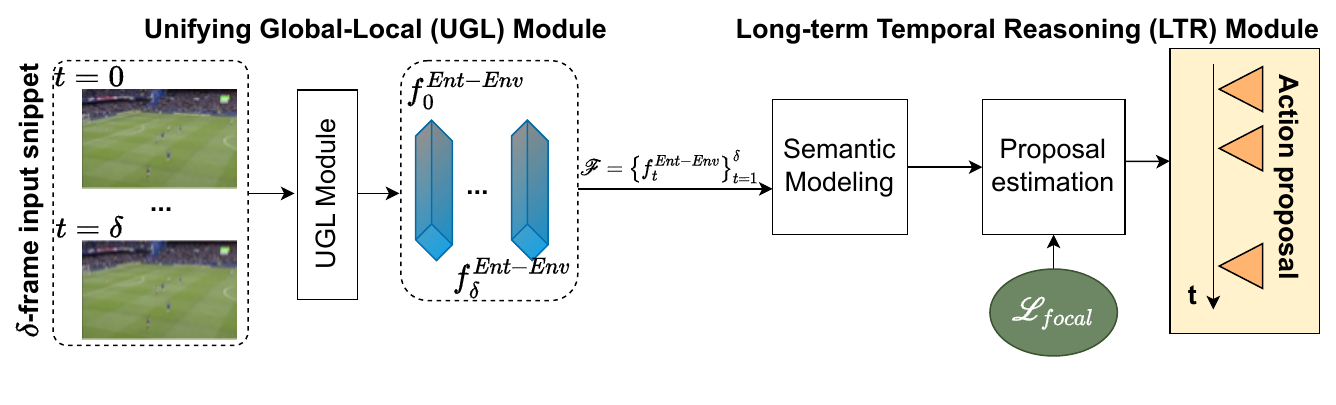}
    \caption{The overall architecture of our proposed network, consisting of Unifying Global-Local (UGL) module and Long-term Temporal Reasoning (LTR) module.}
  \label{fig:full-flow}
\end{figure*}

LTR is responsible for estimating an action score for each frame in the $\delta$-frame snippet. LTR takes $\mathcal{F} = \{f^{Ent-Env}_t\}_{t=1}^{\delta}$ from UGL module as its input. Our LTR contains two components: semantic modeling, and proposal estimation as illustrated in Fig.~\ref{fig:full-flow}. The first component focuses on modeling the semantic relationships between frames within a snippet. This is achieved by applying a 1-layer bidirectional Gated Recurrent Unit (GRU)~\cite{empirical-gru}. The proposal estimation component is implemented using a fully connected (FC) layer and softmax applied to the GRU outputs. This enables the model to make a per-frame prediction in a ${K} + 1$ way classification.

% The proposal estimation component evaluates every frame $v_t$ in the snippet to estimate its actionness score $\hat{\mathbf{y}}_t$. The semantic modeling component is implemented by a 1-layer bidirectional Gated Recurrent Unit (GRU ~\cite{empirical-gru}) network which processes the per-frame G-L features in $\mathcal{F}$. We also experiment with Transformer \cite{attention-is-all-you-need} replacing GRU but do not improve accuracy (see \ref{tab:ablation-test-set-by-class}). The proposal estimation applies a fully connected layer and softmax on the GRU outputs to make a per-frame ${K} + 1$ way prediction.
 
% \begin{figure*}
%     \centering
%     \includegraphics[width=1\linewidth]{images/our-model.pdf}
%     \caption{Our model.}
%     \label{fig:our-model}
%   \label{fig:comparison-models}
% \end{figure*}

At the inference stage, we traverse video through processing ${\delta}$-frame snippets by using the sliding window mechanism with overlapping ${\delta}/2$. Based on the timestamps and scores of candidate proposals, we apply non-maximum suppression (NMS) to produce the set of action proposals.

% \begin{figure*}
%   \centering
%   \begin{subfigure}{1\linewidth}
%     \centering
%     \includegraphics[width=1\linewidth]{images/stanford-model.pdf}
%     \caption{Stanford's model.}
%     \label{fig:stanford-model}
%   \end{subfigure}
%   \hfill
%   \begin{subfigure}{1\linewidth}
%     \centering
%     \includegraphics[width=1\linewidth]{images/our-model.pdf}
%     \caption{Our model.}
%     \label{fig:our-model}
%   \end{subfigure}
%   \caption{Comparison between Stanford's and our model.}
%   \label{fig:comparison-models}
% \end{figure*}

\subsection{Training Methodology}
\label{subsec:training-method}

Upon analyzing the class distribution, for example in SoccerNet-v2 \cite{soccernet-v2}, it is evident that there exists an issue of class imbalance. To be more specific, the classes for red cards and yellow cards have notably fewer samples compared to other classes. Additionally, the majority of frames are categorized as having no events, constituting the background class. More specifically, the ratio between the non-background classes and the background class is approximately 2\%. Due to these factors, it is clear that the imbalanced class distribution poses a significant challenge for model training. Therefore, we propose the adoption of focal loss \cite{focal-loss} as a solution to address this problem.

Particularly, instead of using the cross-entropy as previous methods \cite{yahoo-soares-temporal, e2e_spotting, baidu-feature-combination-meet-attention}, we adopt the focal loss \cite{focal-loss} in \cref{eq:focal-loss}. Focal loss has a modulating factor $(1-y^*_{i,j})^\gamma$ to reshape the loss respective to the difficulty of the samples. Specifically, the loss is down-weighted when it is dealing with the easy sample, while the hard sample penalizes the model strongly when it predicts incorrectly. Moreover, the focal loss also has another parameter $\alpha$ to control the balance between classes.

Given a snippet with $\delta$ frames $v_1,\cdots,v_{\delta}$, for each frame $v_t$, our model outputs an one-hot encoded vector $\hat{\mathbf{y}}_t$ with $K+1$ dimensions. When considering the entire input snippet, our model produces the following output:
$\hat{\mathbf{Y}}=[\hat{\mathbf{y}}_1,\cdots,\hat{\mathbf{y}}_{\delta}]$. With the provided ground-truth $\mathbf{Y}=[\mathbf{y}_1, \cdots, \mathbf{y}_{\delta}]$, the focal loss of the snippet can be defined as follows \cref{eq:focal-loss}:
\begin{equation}
    \begin{array}{rcl}
        \mathcal{L}(\hat{\mathbf{Y}},\mathbf{Y}) & = & \frac{1}{\delta}\sum_{t=1}^{\delta} FL(\bf{\hat{y}}_t, \bf{y}_t) \\
        & = & \frac{1}{\delta}\sum_{t=1}^{\delta}\sum_{k=1}^{K+1} FL(\hat{y}_{t,k}, y_{t,k})
    \end{array}
    \label{eq:focal-loss}
\end{equation}
where $\hat{y}_{t,k}$ and $y_{t,k}$ are the prediction score and the ground-truth score for the $k$-th class of $t$-th frame, respectively. We compute the focal loss for a pair $(\hat{y}_{t,k}, y_{t,k})$ in following \cref{eq:sub-focal-loss}:
% \begin{equation}
%     CE(\hat{\mathbf{y}}_i, \textbf{y}_i) = -\frac{1}{K}\sum_{j=1}^{K+1}{y_{i,j}\log{\hat{y}_{i,j}}}
%     \label{eq:cross-entropy-loss-vector}
% \end{equation}
\begin{equation}
     FL(\hat{y}_{t,k}, y_{t,k}) = \alpha^*_{t,k}[1-y_{t,k}^*]^\gamma BCE(\hat{y}_{t,k},y_{t,k})
     \label{eq:sub-focal-loss}
\end{equation}
where:
% \begin{align}
%     \alpha^*_{t,k} & = \alpha y_{t,k} + (1-\alpha)y_{t,k} \label{eq:fl-all} \\
%     y^*_{t,k} & = \hat{y}_{t,k} y_{t,k} + (1-\hat{y}_{t,k}) (1-y_{t,k}) \label{eq} \\
%     BCE(\hat{y}_{t,k},y_{t,k}) & = y_{t,k} \log(\hat{y}_{t,k}) \label{eq:bce} \\
%     & + (1-y_{t,k}) \log(1-\hat{y}_{t,k}) \notag
% \end{align}
\begin{align}
    \alpha^*_{t,k} & = \alpha \cdot y_{t,k} + (1-\alpha) \cdot y_{t,k} \label{eq:fl-all} \\
    y^*_{t,k} & = \hat{y}_{t,k} \cdot y_{t,k} + (1-\hat{y}_{t,k}) \cdot (1-y_{t,k}) \label{eq}
\end{align}
\begin{equation}
\begin{aligned}
     BCE(\hat{y}_{t,k},y_{t,k}) &=  y_{t,k}\log(\hat{y}_{t,k}) \\
     & + (1-y_{t,k})\log(1-\hat{y}_{t,k})
    \label{eq:bce}
\end{aligned}
\end{equation}

At first, we class-wise calculate the binary cross-entropy (BCE) as \cref{eq:bce}, After that, it is modified by multiplying a difficult control factor $(1-y^*_{i,j})^\gamma$, which controls the sample's difficulty, and a balanced factor ${\alpha}^*$ to obtain the focal loss in \cref{eq:sub-focal-loss}.

%------------------------------------------------------------------------
\section{Experiments}

\subsection{Sports Video Datasets}
\label{sec:dataset}

\begin{figure}[t]
    \centering
    \includegraphics[width=1\linewidth]{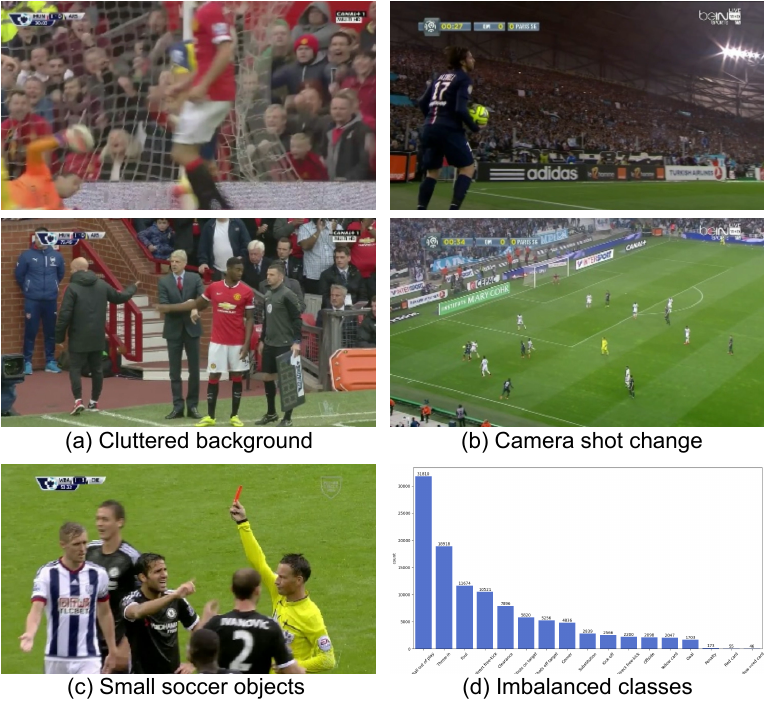}
    \caption{Visualization of four challenges in SoccerNet-v2 data}
    \label{fig:challenges}
\end{figure}

\noindent 
\textit{SoccerNet-v2}\cite{soccernet-v2} consists of 550 soccer matches (1,100 videos) divided into four splits including 600 videos for train, 200 videos for val, 200 videos for test, and 100 videos for challenge set. The video length of the dataset is roughly 764 hours, with over 300,000 labeled time-spot. We also present visualizations of certain scenes to demonstrate the high level of difficulty in SoccerNet-v2 for the action-spotting task in \cref{fig:challenges} such as cluttered background, camera shot change, small soccer objects, and imbalanced classes.

\noindent 
\textit{FineDiving}\cite{finediving} comprises 3,000 diving clips annotated with temporal segments and was relabeled by spotting the step transition frames for four classes, which include transitions into somersaults (pike and tuck), twists, and entry.

\noindent 
\textit{FineGym}\cite{Shao2020FineGymAH} contains 5,374 gymnastics performances, each treated as an untrimmed video, with 32 spotting classes. The original annotations denote the start and end of actions, and E2E-spot\cite{Shao2020FineGymAH} relabeled them, for instance, two classes ``balance beam dismount start'' and ``balance beam dismount end'' were converted from the segments of ``balance beam dismount''.

\subsection{Evaluation metrics}
In the action-spotting task, we use the metrics mAP and tight Average-mAP (T-mAP) to evaluate our proposed network as many benchmarks as well as compare it with SOTA approaches.

Given an acceptable tolerance $\Delta$, a prediction is considered a true positive (TP) if the temporal distance between the prediction and the ground-truth does not exceed the given tolerance.
For each class, the Average Precision (AP) is calculated using the Precision-Recall (PR) curve. Then, the mean-AP (mAP) is the average AP of all the classes. By evaluating the area under the curve (AUC) of the mAP for different tolerances $\Delta$, the result is the Average-mAP (A-mAP).

Previously, SoccerNet-v1\cite{soccernet-v1} evaluated action-spotting models on loose tolerance from 5 seconds to 60 seconds, which is called loose A-mAP. However, for a more accurate evaluation, a new metric - the tight A-mAP (T-mAP) was introduced in SoccerNet-v2\cite{soccernet-v2} with tight tolerance just ranging from 1 second to 5 seconds.

\subsection{Implementation Details}

Within the UGL Module, for the global environment feature extraction component, we utilize the RegNet-Y architecture and initialize it with pre-trained weights sourced from ImageNet-1K \cite{imagenet-ILSVRC15}. On the other hand, for the local relevant scene entities extraction component, we employ a pre-trained model known as GLIP-L \cite{glip1}, which is built upon the Swin-Large backbone.
%Afterward, we freeze GLIP model and use it to do the inference frame by frame, producing the localization of sports objects. By using the RoiAlign technique, we extract object features from the detected objects. Before feeding video frames into the GLIP, we construct a vocabulary list of sports objects that relate to the corresponding sports datasets. These words are then passed into the model to find which objects need to be captured. 

Our UGL is trained by randomly sampling 100-frame-long snippets (snippet length $\delta=100$) and applying standard data-augmentation techniques such as cropping, jittering, and mixup \cite{mixup}. The frames are resized to a height of 224 pixels without cropping for SoccerNet-v2. On FineDiving and FineGym, we randomly crop to 224 $\times$ 224 pixels. We also use the AdamW optimization \cite{adamw}, linear warm-up and cosine LR-annealing \cite{lr-anneling-convex-prob, sgdr-cosine-anneling} with $10^{-4}$ learning rate in our training process. We trained on a single GPU NVIDIA A100-SXM4-80GB with 150 epochs for SoccerNet-v2 and FineGym, and 50 epochs for FineDiving. We use non-max suppression (NMS) to obtain the final time-spots.

Following the best setting of the SOTA methods \cite{yahoo-soares-temporal, e2e_spotting} on SoccerNet-v2, our best model is trained on the combination of train, validation, and test splits, and then evaluated on the challenge set of SoccerNet-v2 and gains top-1 rank on the Leaderboard 
\url{https://eval.ai/web/challenges/challenge-page/1536/leaderboard/3825}.

%  \begin{figure}[t]
%     \centering
%     \includegraphics[width=.8\linewidth]{images/challenge.pdf}
%     \caption{Leaderboard of SoccerNet Challenge 2022 - Action Spotting captured on January 15, 2024.}
%     \label{fig:leaderboard}
% \end{figure}

\subsection{Comparison with the SOTA Methods}
We compare the network with the existing methods on the SoccerNet-v2 challenge set, and the results are shown in \cref{tab:experiment-result-soccer}. Our method achieves 69.38\% T-mAP and shows the best performance, surpassing other SOTA methods \cite{yahoo-soares-temporal, e2e_spotting}. Compared to E2E-spot \cite{e2e_spotting}, which has the same training setting as ours, our network outperforms all metrics, especially achieving a notable gap of 2.65\% T-mAP. Our work also outperforms two-phase approaches Baidu \cite{baidu-feature-combination-meet-attention} (1st place in 2021) and Soares \cite{yahoo-soares-temporal} (1st place in 2022), by 19.82\% and 1.57\% T-mAP, respectively. Our network shows strong performance on shown actions with a significant gap of 2.66\% T-mAP compared to Soares \cite{yahoo-soares-temporal}, while Soares still keeps the best result on unshown actions (not visible in the frame).

\cref{tab:experiment-result-sports} shows the comparison between our method with baselines and SOTA methods on the test set of FineDiving \cite{finediving} and FineGym \cite{Shao2020FineGymAH} datasets. Following E2E-Spot\cite{e2e_spotting}, we perform a comprehensive benchmarking of the system across three distinct scenarios. (a): Pre-trained features (from Kinetics-400\cite{Kay2017TheKH}), (b): Fine-tuned features, (c): VC-Spot: video classification baseline using RGB. Our network excels in performance across all metrics on both datasets. Within tolerance $\Delta=2$, we surpass the baselines including (a), (b), and (c) by an average of 12.03\%, and 15.3\% mAP on FineDiving, and FineGym datasets, respectively. For a fair comparison with the SOTA method, we compare UGL with E2E-Spot (black) without using flow\cite{Teed2020RAFTRA}, UGL outperforms E2E-Spot by 2.4\%, and 1.3\% mAP at $\Delta=2$ on FineDiving, and FineGym datasets, respectively.

% \begin{table}
%     \centering
%     \resizebox{\linewidth}{!}{
%     \begin{tabular}{ l | c c | c c }
%         \multirow[ht]{2}{*}{Method} & 
%         \multicolumn{2}{c|}{Test set} & 
%         \multicolumn{2}{c}{Challenge set} \\
%         & Tight & Loose & Tight & Loose \\
%         \hline
%         CALF \cite{calf} & - & - & 15.33 & 42.22 \\
%         CALF-calib \cite{calf-calib} & - & 46.80 & 15.83 & 46.39 \\
%         RMS-net \cite{rms-net} & 28.83 & 63.49 & 27.69 & 60.92 \\
%         NetVLAD++ \cite{netvladpp} & - & - & 43.99 & 74.63 \\
%         Zhou et al. \cite{baidu-feature-combination-meet-attention} & 47.05 & 73.77 & 49.56 & 74.84\\
%         Soares et al. \cite{yahoo-soares-temporal} & \textbf{65.07} & \textbf{78.59} & $67.81^*$ & $\textbf{78.05}^*$ \\
%         \hline
%         E2E-Spot (base) \cite{e2e_spotting} & 61.82 & 74.05 & $66.73^*$ & $73.26^*$  \\
%         UGLF-Net (ours) & 62.49 & 73.98 & $\textbf{69.38}^*$ & $76.14^*$
%     \end{tabular}}
%     \caption{Performance comparisons on both test and challenge sets of SoccerNet-v2 in terms of tight A-mAP and loose A-mAP. * indicates for models trained on the combination of the train, validate, and test splits. \textbf{Bold} is the highest value. }
%     \label{tab:experiment-result}
% \end{table}

\begin{table}
    \centering
    \caption{Performance comparisons on SoccerNet-v2 challenge set in terms of tight Avg-mAP (T-mAP). Shown and unshown refer to whether actions are visible. \textbf{Bold} is the highest value.}
    \resizebox{\linewidth}{!}{
    \begin{tabular}{ l | c | c c }
    \toprule
        \textbf{Method} & \textbf{T-mAP} & \textbf{Shown} &  \textbf{Unshown} \\ \hline
        CALF \cite{calf} & 15.33 & - & - \\
        RMS-net \cite{rms-net} & 27.69 & - & - \\
        NetVLAD++\cite{netvladpp} & 43.99 & - & - \\
        CALF-calib \cite{calf-calib} & 46.80 & - & - \\        
        Baidu ($1^{st}$ 2021) \cite{baidu-feature-combination-meet-attention} & 49.56 & 54.42 & 45.42 \\
        Transformer-AS\cite{Zhu2022ATS} & 52.04 & 60.18 & 32.06 \\
        Faster-TAD\cite{Chen2022FasterTADTT} & 64.88 & 70.31 &	53.03 \\
        E2E-Spot ($2^{nd}$ 2022) \cite{e2e_spotting} & $66.73$ & $74.84$ & $53.21$  \\        
        Soares ($1^{st}$ 2022) \cite{yahoo-soares-temporal} &  $67.81$ & $72.84$ & $\textbf{60.17}$ \\
        \hline
        \textbf{Ours} & $\textbf{69.38}$ & $\textbf{75.50}$ & $53.31$ \\
    \bottomrule
    \end{tabular}}
    \label{tab:experiment-result-soccer}
\end{table}

\begin{table}
    \centering
    \caption{Spotting performance (mAP @ $\Delta$ frames). The top results are \textbf{bold}. Symbol † indicates NMS. E2E-Spot (gray) uses 2-streams input (ensemble with flow \cite{Teed2020RAFTRA}), and E2E-Spot (black) uses only RGB input.}
    \resizebox{\linewidth}{!}{
    \begin{tabular}{ ll|cc|cc }
    \toprule    
        \multirow[ht]{2}{*}{\textbf{Features}} & \multirow[ht]{2}{*}{\shortstack{\textbf{Temporal}\\\textbf{Encoder}}} & \multicolumn{2}{c}{\textbf{FineDiving}\cite{finediving}} &  \multicolumn{2}{c}{\textbf{FineGym}\cite{Shao2020FineGymAH}} \\ 
        & & $\Delta$=1 & $\Delta$=2 & $\Delta$=1 & $\Delta$=2 \\ \hline
        \multicolumn{4}{l}{\emph{(a) Pre-trained features (from Kinetics-400)}} \\
         MViT-B\cite{Fan2021MultiscaleVT} & MS-TCN\cite{AbuFarha2019MSTCNMT} & 59.3 & 78.3$^\dag$  & 31.0$^\dag$ & 48.6$^\dag$ \\
         MViT-B\cite{Fan2021MultiscaleVT} & GRU\cite{empirical-gru}  &   57.3 & 76.7  & 28.5$^\dag$ & 48.6$^\dag$ \\
         MViT-B\cite{Fan2021MultiscaleVT} & AsFormer\cite{chinayi_ASformer} &  38.5 & 67.4$^\dag$  & 25.3$^\dag$ & 42.9$^\dag$ \\ \hline
         \multicolumn{4}{l}{\emph{(b) Fine-tuned features}} \\
         TSP\cite{Alwassel2020TSPTP}  &  MS-TCN\cite{AbuFarha2019MSTCNMT} &  57.7 & 76.0$^\dag$  & 40.5$^\dag$ & 58.5$^\dag$ \\
         TSP\cite{Alwassel2020TSPTP} & GRU\cite{empirical-gru} & 57.0 & 78.2  & 38.7$^\dag$ & 58.8$^\dag$ \\  
         TSP\cite{Alwassel2020TSPTP} & ASFormer\cite{chinayi_ASformer} &  51.3 & 77.4$^\dag$  & 38.8$^\dag$ & 57.6$^\dag$ \\ \hline
         \multicolumn{4}{l}{\emph{(c) VC-Spot: video classification baseline using RGB}} \\
         RegNet-Y\cite{regnet-y} & VC-Spot & 62.4$^\dag$ & 85.6$^\dag$  & 18.7$^\dag$ & 28.6$^\dag$ \\ \hline
         \multicolumn{4}{l}{\emph{(d) E2E-Spot}} \\
         RegNet-Y\cite{regnet-y} & GRU\cite{empirical-gru} & 68.4 & 85.3$^\dag$  & 49.0$^\dag$ & 66.5$^\dag$ \\
         \color{gray}{RegNet-Y\cite{regnet-y}} & \color{gray}{GRU\cite{empirical-gru}} & \color{gray}{66.4$^\dag$} & \color{gray}{84.8$^\dag$}  & \color{gray}{51.8$^\dag$} & \color{gray}{68.5$^\dag$} \\      
         \hline 
         \textbf{Ours} & GRU\cite{empirical-gru} & \textbf{70.0} & \textbf{87.7$^\dag$} & \textbf{50.2$^\dag$} & \textbf{67.8$^\dag$} \\
    \bottomrule
    \end{tabular}}
    \label{tab:experiment-result-sports}
\end{table}

\subsection{Ablation Studies}
Within this section, we carry out ablation studies to assess the influence of each component within our network. This encompasses the examination of various types of features, temporal reasoning mechanisms, and loss functions. Additionally, we place emphasis on the effectiveness of our network concerning several long-tail classes as shown in \cref{tab:ablation-test-set-by-class}. In this table, we consider two variations for the temporal encoder: GRU (our choice) and Transformer Encoder, along with two options for the loss function: Cross-Entropy (CE) and Focal Loss (FL). This ablation study places particular focus on long-tail classes such as ``penalty'', ``red card'', and ``yellow to red card''.

% In this section, we study the components of our method, including kinds of extracted features, temporal reasoning, and loss function, and also show the efficiency of our UGLF-Net on several long-tail classes in \cref{tab:ablation-test-set-by-class}.

% \begin{table}
%     \centering
%     \caption{Ablation study of kinds of extracted feature, temporal reasoning, loss function on the SoccerNet-v2 test set with tight A-mAP. \textbf{Bold} is the highest value.}    
%     \resizebox{1.0\linewidth}{!}{
%     \begin{tabular}{l|ccc|c}
%         Method & \makecell{Yellow} & \makecell{Red} & \makecell{Yellow \\to red} & \makecell{Tight\\A-mAP}\\
%         \hline
%         \makecell{Global + GRU + CE\\(E2E-Spot \cite{e2e_spotting})} & \textbf{64.69} & 18.87 & 17.1 & 61.82 \\
%         \hline
%         Global + TE1 + CE & 56.27 & 3.69 & 2.96 & 52.05 \\
%         Global + TE3 + CE & 56.90 & 7.25 & 2.07 & 53.97 \\
%         \hline
%         Global + GRU + FN & 60.91 & 12.43 & 5.66 & 60.13 \\
%         Global + GRU + FH & 62.39 & \textbf{28.97} & 11.04 & 61.92 \\
%         \hline
%         Fusion + GRU + CE & 58.73 & 28.43 & 14.38 & 57.21\\
%         Fusion + GRU + FH & 55.90 & 27.57 & \textbf{20.36} & 58.35\\
%         \hline
%         UGLF-Net & \textbf{64.69} & 27.57 & \textbf{20.36} & \textbf{62.50}
%     \end{tabular}
%     }
%     \label{tab:ablation-test-set}
% \end{table}

\begin{table}
    \centering
    \setlength{\tabcolsep}{2pt}
    \renewcommand{\arraystretch}{1.0}
    \caption{Ablation study of kinds of extracted feature, temporal reasoning, loss function on several long-tail classes (Penalty, Red card, Yellow to red) on the SoccerNet-v2 test set with tight A-mAP. CE, FL, GRU denote Cross-Entropy,  Focal Loss and Gated Recurrent Unit. \textbf{Bold} is the best value.}
    \resizebox{1.0\linewidth}{!}{
    \begin{tabular}{l|l|l||llll}
    \toprule
    \multirow{2}{*}{\textbf{Features}} & \textbf{Temporal} & \multirow{2}{*}{\textbf{Loss}} & \multirow{2}{*}{\textbf{Penalty}} & \multirow{2}{*}{\textbf{Red}} & \multirow{2}{*}{\textbf{Yell.2Red}} & \multirow{2}{*}{\textbf{T-mAP}} \\ 
    & \textbf{Encoder} & & & & \\ \hline
    {$f^{Env}$( Global }   & {Transformer}                & CE   & 82.92       & 7.25   & 2.07             & 53.97   \\ \cline{3-7}
    Environment)         &           Encoder        & FL   &   83.07     & 3.69  & 2.37             &  52.05  \\ \cmidrule{1-7}
     {$f^{Env}$(Global}         &  \multirow{2}{*}{GRU}               & CE   & 79.82       & 28.61   & 5.86         & 61.82   \\ \cline{3-7}
     Environment)        &                   & FL   & 83.50       & 28.97   & 11.04         & 61.92   \\ \cmidrule{1-7}
    $f^{Ent-Env}$    &  \multirow{2}{*}{GRU}               & CE   & 84.29       & 35.73              & 14.38         & 62.43   \\ \cline{3-7}
     (Fusion)        &                   & FL   & \textbf{86.73}       & \textbf{36.90}   & \textbf{20.36}         & \textbf{63.51}  \\ \bottomrule
    \end{tabular}
    }
    \label{tab:ablation-test-set-by-class}
\end{table}

\begin{table}
    \centering
    \setlength{\tabcolsep}{9pt}
    \caption{Ablation study of $\gamma$ in Equation \ref{eq:sub-focal-loss} on SoccerNet-v2\cite{soccernet-v2} test set. \textbf{Bold} is the best value.}    
    \resizebox{1.0\linewidth}{!}{
        \begin{tabular}{l|llll}
        \toprule
        $\gamma$          & \textbf{Penalty} & \textbf{Red} & \textbf{Yell2Red} & \textbf{T-mAP} \\ \hline
        2           &    83.62     &  35.10   &      12.27    &   62.40  \\ \hline
        4           &     85.57    &  36.01   &   18.43       &   62.94  \\ \hline
        5 (default)          &  \textbf{86.73} & \textbf{36.90} & \textbf{20.36} & \textbf{63.51}   \\ \hline
        6           &    86.12     &  36.47   &   20.15       &  63.12   \\ 
        \bottomrule
        \end{tabular} 
    }
    \label{tab:ablation-gamma}
\end{table}

When utilizing the same feature, specifically the global environment feature $f^{Env}$, it becomes evident that GRU outperforms the Transformer Encoder in both cases, whether the loss function is CE or FL. Additionally, when considering the same feature selection and temporal encoder, FL demonstrates clear advantages compared to CE. Furthermore, when using the same temporal encoder and loss function, it is apparent that the fused feature $f^{Ent-Env}$, which combines both the global environment feature and local relevant scene entities, outperforms the network that relies solely on the global feature.
% From \cref{tab:ablation-test-set-by-class}, the Transformer does not work very well. As we have mentioned before, the purpose of putting focal loss in our model is to mitigate the imbalance problem because it can create a strong penalty on the samples of long-tail classes.

% By replacing the global feature model with our UGLF feature, the performance of long-tail classes "penalty", "red card" and "yellow to red card" rise significantly. 

\cref{tab:ablation-gamma} provides an ablation study on the hyper-param $\gamma$ defined in Focal Loss, Equation \ref{eq:sub-focal-loss}. In our implementation, we set $\gamma$ as 5.

% Combining with the previous insight, we change the loss function to the high gamma focal loss, as a result, the performance on "red card" and "yellow to red card" have increased from 18.87 A-mAP to 27.57 A-mAP (for red card class), and from 17.1 A-mAP to 20.36 A-mAP (for yellow to red card class).

% \begin{table}
%     \centering
%     \caption{Ablation study on several specific classes (Direct free-kick, Red card, Yellow to red) on the SoccerNet-v2 test set between the SOTA E2E-spot\cite{e2e_spotting} and our proposed method. \textbf{Bold} is the highest value.}    
%     \resizebox{1.0\linewidth}{!}{
%     \begin{tabular}{ l | c c c | c }
%         Method & \makecell{Direct\\free-kick} & \makecell{Red\\card} & \makecell{Yellow\\to red} & \makecell{Tight\\A-mAP}\\
%         \hline
%         E2E-Spot & 69.20 & 18.87 & 17.10 & 61.82 \\
%         UGLF-Net (ours)& \textbf{71.82} & \textbf{27.59} & \textbf{20.36} & \textbf{62.49}
%     \end{tabular}
%     }
%     \label{tab:ablation-test-set-by-class}
% \end{table}

%------------------------------------------------------------------------
\section{Conclusion}
In this study, we introduce an effective framework designed for the task of action spotting within sports videos. Our model follows an end-to-end approach, wherein we enhance visual feature representation by incorporating a pre-trained vision-language model and Adaptive Attention Mechanism to extract locally relevant scene entity features. Consequently, our network considers both the global environment feature and the local relevant scene entities feature for action spotting. To address challenges related to imbalanced distribution and long-tail classes, we employ Focal Loss during training. Our experiments demonstrate that our network outperforms existing methods across multiple datasets.

Furthermore, most of the existing action spotting methods are designed as a black-box, making them challenging to interpret their results. Conversely, our proposed network provides interpretability, allowing it to explain which elements of the scene contribute to the action and explain the relationships between scene entities and their surrounding environment.
% After that, we combine the local features and global features (extracted from CNN-2D backbone) into the global-local spatial features. The proposed model achieves the top-1 rank on the SoccerNet-v2 action-spotting challenge. In addition, we evaluate the model performance by class to compare it with the E2E-spot model. Experimental results and ablation studies have shown the effectiveness and potential of adding prior (language) knowledge to the video understanding model for action-spotting task.

% Most of the existing models for action-spotting in soccer videos are black boxes which makes it difficult to explain the result of the model. By adding the local features of related-soccer objects, our proposed idea also helps the outcome to be more explainable compared to other deep learning models.

\section{Future Work}

In the future, we employ a Graph Neural Network (GNN) to model the relationships between relevant scene entities. In this framework, scene entities are treated as nodes, and the relationship between them is represented as edges. This approach allows us to incorporate prior knowledge about the rules of the sport into the model, thereby enhancing both the model's learning capabilities and its explainability.

Finally, to address the imbalanced class problem, grouping related classes into groups is essential, for instance, the classes ``yellow card'', ``red card'', and ``yellow to red'' are grouped into a ``cards'' group, and each spot as a candidate is classified into a group before determining the final action.

\bibliographystyle{abbrv}
\bibliography{ref}

\end{document}